\documentclass[letterpaper, 10 pt, conference]{ieeeconf}

\IEEEoverridecommandlockouts                              

\usepackage{xcolor}
\usepackage{amsmath} 
\usepackage{amssymb}  
\usepackage{graphicx}
\usepackage{cite}

\usepackage{fancyhdr}
\fancypagestyle{firstpage}{
    \chead{This paper has been accepted for publication at the 2026 IEEE/RSJ International Conference on Intelligent Robots and Systems (IROS 2026)}}

\usepackage{url}

\usepackage{algorithm}      
\usepackage{algpseudocode}  

\newtheorem{definition}{Definition}

\newtheorem{problem}{Problem}

\newcommand{\argmin}{\mathop{\mathrm{argmin}}\limits}

\title{\LARGE \bf
An Automated Thickness Evaluation Procedure Using an Integrated Structured Light 3D Camera in a Robotic Bioprinting Framework
}

\author{Ehsan Zobeidi$^{*1}$, Omid Rezayof$^{*1}$, Farshid Alambeigi$^{1}$
\thanks{* These authors contributed equally to this work.}
\thanks{© 2026 IEEE. Personal use of this material is permitted. Permission from IEEE must be obtained for all other uses, in any current or future media, including reprinting/republishing this material for advertising or promotional purposes, creating new collective works, for resale or redistribution to servers or lists, or reuse of any copyrighted component of this work in other works.}
\thanks{Research reported in this
publication was supported by the National Institute Of Arthritis And Musculoskeletal And Skin Diseases of
the National Institutes of Health under Award Numbers DP2AR082471.}
\thanks{$^{1}$E. Zobeidi, O.~Rezayof, and F.~Alambeigi are with the Walker Department of Mechanical Engineering and Texas Robotics, University of Texas at Austin, TX, USA.
Email: {\tt\small \{ehsan.zobeidi, omid.rezayof\}@utexas.edu}, and \tt\small farshid.alambeigi@austin.utexas.edu}.
}

\begin{document}

\maketitle
\thispagestyle{firstpage}
\pagestyle{empty}

\begin{abstract}
Bioprinting is emerging as a tissue engineering technique to replace common treatment methods for large scale injuries.
While thickness of the BioPrinted Constructs (BPCs) have shown to be of importance in the cell maturation and integration, the literature lacks a robust, automated, and quantitative method for measuring these metrics. In this paper, we propose a fully automated vision-based method for measuring the thickness of the BPCs with complex geometries.
Leveraging the point cloud and RGB images of a structured light 3D camera, our proposed method performs an image-based segmentation for delineating the BPCs from the RGB images, accompanied by novel geometry-based thickness measurement algorithms performed on the point cloud scans.
These algorithms combine the segmentation mask with the robot's forward kinematics data and a 3D point cloud scan to precisely measure the aforementioned metrics for complex-shaped BPCs.
The proposed method was evaluated in simulation and experimental studies.
In simulation studies, the algorithms were used to measure the thickness of some virtually created BPCs with known thickness.
The comparison between the measured and true thicknesses demonstrates the high accuracy of the proposed method, achieving mean absolute errors between $0.025$\,mm and $0.057$\,mm in simulation at a spatial resolution of $0.1\,\mathrm{mm}\times0.1\,\mathrm{mm}$ per pixel.
Furthermore, we successfully deployed the algorithms on our robotic bioprinting setup utilizing a structure light 3D camera, where complex patterns were printed and the developed methods  utilized to accurately measure the thickness of printed BPCs.

\end{abstract}



\section{Introduction}

Major skeletal muscle injuries, such as those resulted from Volumetric Muscle Loss (VML) injuries, often exceed the tissue’s natural ability to repair itself. Consequently, clinical intervention is required to restore muscle function and prevent permanent physical impairment \cite{carnes2020skeletal}.
The established clinical treatment for VML involves surgical grafting of healthy tissue, which is limited by tissue availability and poor integration between the implanted graft and the surrounding native tissue \cite{Jana2016AnisotropicMF}.
Recently, bioprinting is emerging as a promising approach to address these limitations in VML injury treatment \cite{ashammakhi2019situ}.
In vitro bioprinting enables the fabrication of complex tissue constructs in controlled laboratory environments; however, despite significant progress (e.g., \cite{Kang2016A3B}), its clinical applicability remains limited due to poor tissue integration and challenges associated with implanting BioPrinted Constructs (BPCs) into defect sites \cite{Manyi}.
In contrast, in situ bioprinting directly deposits biomaterials into the damaged site, offering improved integration with native tissue \cite{carnes2020skeletal, singh2020situ}.
\begin{figure}
\includegraphics[width=\columnwidth]{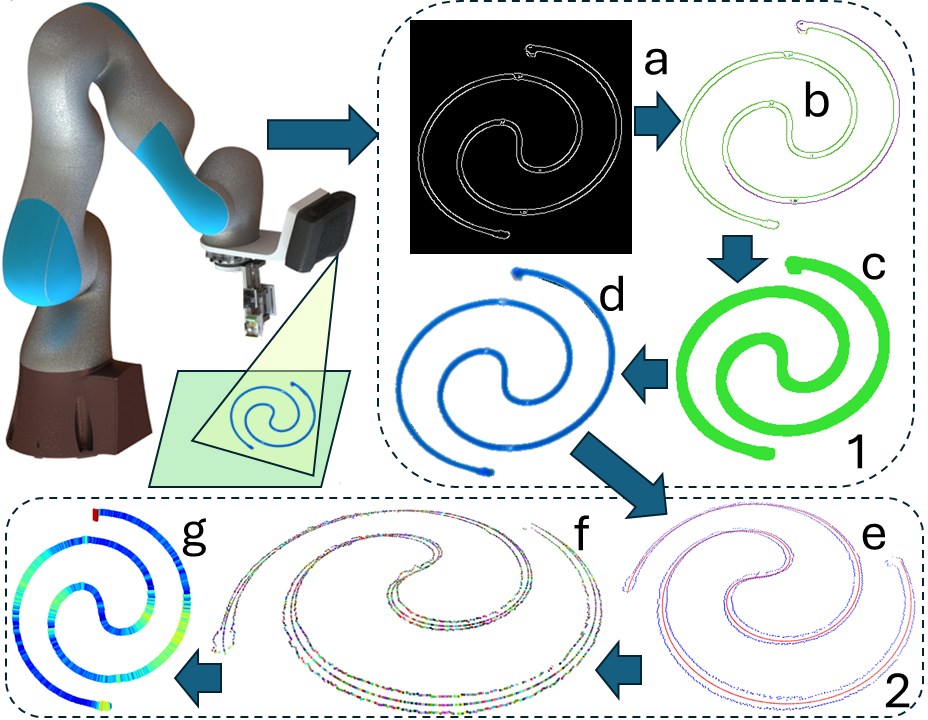}
\caption{This figure illustrates the steps of our method. After the robot deposits the bio-printed material, RGB and 3D point cloud images are captured. Step 1 (Sec.~\ref{subsec:boundary}) performs delineation and segmentation of the BPC using the RGB image. Boundary pixels are extracted using the Canny detector (\textsf{a}). Raw contours (filled) are shown in \textsf{b}, but these may self-overlap and fail to separate the interior from the exterior. Rough contours (\textsf{c}) help distinguish interior from exterior but are imprecise. The final precise contour is obtained by fusing raw and rough contours (\textsf{d}).
Step 2 (Sec.~\ref{subsec:width}) calculates thickness. The 3D points include robot deposition path points (red) and fused contour vertices (blue) (\textsf{e}). Side points are computed for each path point (\textsf{f}); each path point and its side points share the same color for visualization. Thickness is measured as the distance between side points and visualized along the path (\textsf{g}).}
\label{fig:concept}
\vspace{-6mm}
\end{figure}

Handheld bioprinters have been explored in prior studies towards in situ bioprinting (e.g., \cite{Nuutila2021InVP, Rezayof2024OnTPISMR}); however, their printing precision is inherently limited by manual operation. Consequently, robotic bioprinting approaches have been introduced to achieve improved robustness and precision \cite{Albanna2019InSB, ALevin2023CommercialAC, yang2024development, Rezayof2024}. Despite these advances, existing studies largely lack quantitative metrics for evaluating the geometric quality of the resulting BPCs.
For instance, Fortunato et al. \cite{Fortunato2021} employed a 5 Degree-of-Freedom (DoF) robotic system for in situ bioprinting and evaluated the motion accuracy of the platform; however, the resulting BPCs were not directly subjected to quantitative geometric assessment. Similarly, Levin et al. \cite{ALevin2023CommercialAC} utilized a 7 DoF robotic bioprinting system and compared RGB images of the printed constructs with the commanded trajectories, providing an indirect evaluation of motion accuracy rather than the geometry of the BPCs themselves.
Meanwhile, prior studies have demonstrated that the alignment and uniformity of printed constructs significantly influence cell growth and maturation \cite{carnes2020skeletal, Kim20183DBH}. Therefore, fast and reliable quantitative measurements of BPC geometry are essential for optimizing bioprinting parameters, such as deposition rate and robot velocity, to achieve accurate and reproducible construct patterns.

To address this problem, several studies have proposed robotic bioprinting frameworks equipped with complementary quantitative algorithms for evaluating thickness and uniformity \cite{yang2024development, Rezayof2024}. In these works, a point cloud camera is used to scan the BPCs, followed by segmentation to separate the BPCs from the background. Geometry-driven algorithms are then applied to measure their thickness.
Specifically, the approach proposed in \cite{yang2024development, Rezayof2024} employs the GrabCut algorithm for segmentation \cite{rother2004grabcut}, followed by an energy-based refinement method to achieve sub-pixel masking precision \cite{li2010distance}. While this framework represents one of the first efforts to provide quantitative evaluation of the BPCs, the manual selection of input samples limits its full automation capability. In addition, the energy-based boundary refinement may lead to partial loss of information related to potential non-uniformities in the BPCs.
Moreover, the previously proposed thickness measurement methods rely on restrictive geometric assumptions. For example, the approaches developed in \cite{yang2024development} and \cite{Roach2023InvertibleNN} are limited to BPCs printed in straight lines, while the method presented in \cite{Rezayof2024} is limited to curves that can be represented as single-valued mappings from the $x$-dimension to the $y$-dimension. As a result, it fails for geometries that curve back onto themselves, such as spirals or looped structures as shown in Fig.~\ref{fig:concept}.

To address these limitations, in this work, we present a novel approach for fully automatic thickness measurement of BPCs.
Leveraging the point cloud and RGB images of a structured light 3D camera, our approach includes segmentation of BPCs from the background along with complementary geometry-driven algorithms to measure their thickness. Boundary pixels are detected using the Canny edge detector on the RGB image \cite{ding2001canny}, and the resulting contours are processed to delineate the BPC. The 3D BPC thickness is then computed using the point cloud scan, the recorded path of the robot end effector, and the segmentation results.
The proposed method offers two main advantages: (i) it is fully automatic, requiring no user input, and (ii) is computationally efficient, both of which enable future real-time optimization of printing parameters during the printing procedure. As the approach relies on Canny-based contours, the method ensures that boundaries are accurately captured without the need for slow optimization-based methods such as GrabCut.
The performance and effectiveness of the algorithms are validated through both simulation and experimental studies.
In fact, the accuracy and trueness of the measurements are first evaluated in the simulations studies, where the proposed algorithms are used to measure virtual BPCs, carefully designed to have specific thickness values. The true and measured thickness are then compared to evaluate the accuracy of the algorithms. Furthermore, experimental studies are conducted on a robotic bioprinting system, demonstrating the capabilities of the proposed methods on hardware.


\section{Problem Formulation}\label{sec:problem_formulation}

Given an RGB image of the bio-printed material against a background, as illustrated in Fig.~\ref{fig:segmentation}, our goal is to precisely delineate the bio-printed material in the $H \times W$ image, which will be used later for thickness calculation. Of note, in this paper, the term ``\textit{thickness}" refers to the dimension of the printed filaments parallel to the surface and perpendicular to the direction of printing filaments. This dimension should not be confused with the filaments' ``\textit{height}" which is perpendicular to the surface.

\begin{definition}[Contour]
A contour is a directed closed curve. In this paper, a contour is represented as a closed polygon. For a contour $\mathbf{C}$ consisting of vertices ${\bf x}_i \in \mathbb{R}^2$ (or $\mathbb{R}^3$), we represent it as a cyclic list:
\[
\mathbf{C} := [\mathbf{x}_0, \dots, \mathbf{x}_l],
\]
where $\mathbf{x}_i$ is connected to $\mathbf{x}_{i+1}$ with a straight line, and $\mathbf{x}_l$ is connected to $\mathbf{x}_0$.
In the 2D case, for a contour $\mathbf{C}$, we define $I_{\mathbf{C}}$ as the set of pixels inside the contour, i.e., those encompassed by $\mathbf{C}$.
\end{definition}

The pixels of the $H \times W$ image can be divided into two sets: the bio-printed material pixels $\mathcal{B}$ and the background pixels $\mathcal{G}$.

\begin{problem}\label{problem:segmentation}
Given an $H \times W$ RGB image containing bio-printed material (with pixel coordinates in $\mathcal{B}$) and background (with pixel coordinates in $\mathcal{G}$), find the minimum number of contours $\mathbf{C}_0, \dots, \mathbf{C}_m$ such that
\[
\bigcup_{i=0}^m I_{\mathbf{C}_i} = \mathcal{B}.
\]
\end{problem}

\begin{problem}\label{problem:widthcalculation}
Given an $H \times W \times 3$ point cloud (from the 3D scanning camera), where each pixel in the RGB image from Problem~\ref{problem:segmentation} has a corresponding 3D point in the point cloud (possibly noisy), and the deposition needle path
\[
Path = \{\mathbf{p}_0, \dots, \mathbf{p}_n\},
\]
determine the thickness of the bio-printed material along the path.
\end{problem}


\section{Approach}\label{sec:approach}
In this section, we present our proposed solutions to the problems formulated in Sec.~\ref{sec:problem_formulation}.
We first address the segmentation problem in Sec.~\ref{subsec:boundary}, followed by our solution to the thickness calculation problem in Sec.~\ref{subsec:width}.

\subsection{Delineation and Segmentation}\label{subsec:boundary}
In this section, we address Problem~\ref{problem:segmentation}. Given an $H\times W$ RGB image containing the bio-printed material and background, our goal is to delineate the bio-printed material as several contours. Note that the bio-printed material can consist of multiple disconnected regions, as illustrated in Fig.~\ref{fig:segmentation}. For each distinct region, we extract one contour. Precise delineation is crucial for the subsequent thickness (width) calculation, which is used to evaluate the quality of the bio-printing process. Therefore, it is essential to accurately capture jumps, discontinuities, and avoid excessive smoothing.

Our approach employs the Canny edge detector to locate boundary pixels via the image gradient. The pipeline consists of three main steps: (i) extraction of raw contours, (ii) generation of rough contours, and (iii) fusion of raw and rough contours to obtain precise delineation.

\subsubsection{Raw Contours}\label{sec:rawcontours}
First, we apply the Canny detector to identify boundary pixels, using image gradients, and extract raw contours ${\bf{C}}_0^{raw},..., {\bf{C}}_l^{raw}$ using the contour-finding algorithm \cite{suzuki1985topological}.
Although raw contours are closed curves, they often fail to encompass a complete bio-printed region and distinguish interior from exterior (Fig.~\ref{fig:segmentation}). Mostly, the contours are closed in the sense that they pave a part of the boundary and then trace back on themselves. Multiple raw contours may correspond to a single region, and small internal boundaries may appear due to reflections.
Nevertheless, raw contours exhibit two key advantages: (i) their pixels lie precisely on actual boundaries, and (ii) the order of pixels in sub-sequences remains correct, which is critical for reconstructing continuous contours.

\subsubsection{Rough Contours}\label{sec:roughcontours}
\begin{figure*}[t]
\centering
\includegraphics[width=0.9\linewidth]{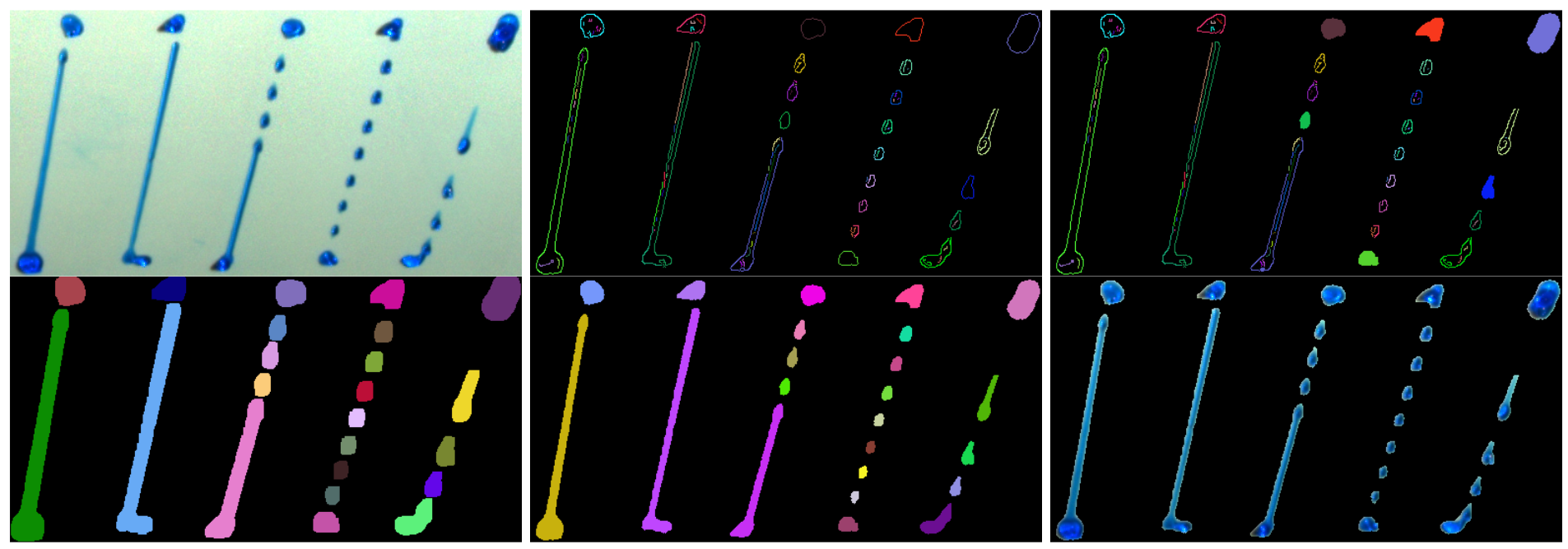}

\caption{On the first row, from left to right, the first image is a sample image of bio-printed material with background. The second image is the raw contours, and the third image is the raw contours, but they are filled. We can see that most of the raw contours encompass a very small space; as a result, they can not determine the inside from the outside of the region. For one region, we also observe that several raw contours are assigned to one region. However, their good quality is that they are on the boundary of the region. On the second row, from left to right, the first image is the rough contours (that is, dilated with a $7\times 7$ kernel), which are filled. We can see that the rough contours are getting the region correctly, but their boundary is not precise, and it is faded and interpolated. The second image is the final and precise contours, which are filled after the fusion between the rough contours and the raw contours. By observing the filled version of fused contours, we see that the resulting contours benefit from both qualities of raw and rough contours, that both of the boundaries are accurate, and each contour encompasses a distinct region. In the third image, we use the final fused contours to extract the bio-printed pixels ($\mathcal{B}$) from the original image. }
\label{fig:segmentation}
\vspace{-5mm}
\end{figure*}

Rough contours prioritize covering entire regions rather than boundary precision. Missing boundary pixels in the raw contours often prevent correct region distinction, highlighting the need for interpolation.
We generate rough contours ${\bf{C}}_0^{rough},..., {\bf{C}}_k^{rough}$ via two steps: (i) we fill the raw contours to interpolate missing boundary pixels, and (ii) we dilate the filled regions \cite{khosravy2017morphological} to close gaps. Applying Canny detection and contour-finding on this interpolated image yields rough contours.
As shown in Fig.~\ref{fig:segmentation}, rough contours cover distinct regions efficiently ($k=m$), i.e., $I_{{\bf{C}}_i} \subset I_{{\bf{C}}_i^{rough}}$, but their boundaries are not precise, often overlapping background pixels.

\subsubsection{Fusion of Raw and Rough Contours}

To obtain the final precise contours ${\bf C}_0,\ldots,{\bf C}_m$, we fuse the raw and rough contours by exploiting their complementary properties: raw contours lie accurately on the true material boundaries and preserve the correct pixel ordering, but do not reliably enclose distinct regions, whereas rough contours provide reliable region separation but suffer from boundary imprecision.

For each rough contour ${\bf C}_i^{\mathrm{rough}}$, we identify the corresponding raw contours ${\bf C}_j^{\mathrm{raw}}$ whose vertices mostly lie inside the region enclosed by ${\bf C}_i^{\mathrm{rough}}$, i.e.,
\begin{equation}\label{eq:correspondingdetection}
    \frac{|I_{{\bf C}_i^{\mathrm{rough}}}\cap {\bf C}_j^{\mathrm{raw}}|}{|{\bf C}_j^{\mathrm{raw}}|} > \frac{1}{2}.
\end{equation}
This criterion assigns a raw contour to a rough contour if more than half of its vertices are located inside the rough region, and significantly reduces the search space in the subsequent steps.
Next, for each rough contour, we extract and stitch the relevant sub-sequences of its corresponding raw contours. Without loss of generality, consider ${\bf C}_0^{\mathrm{rough}}$ with corresponding raw contours ${\bf C}_0^{\mathrm{raw}},\ldots,{\bf C}_p^{\mathrm{raw}}$. For each vertex $[{\bf C}_0^{\mathrm{rough}}]_s$, we select the closest vertex among all corresponding raw contours:
\begin{equation}
    q_s, r_s = \arg\min_{i,j} \left\|[{\bf C}_0^{\mathrm{rough}}]_s - [{\bf C}_i^{\mathrm{raw}}]_j \right\|_2 .
\end{equation}
The selected vertices $[{\bf C}_{q_s}^{\mathrm{raw}}]_{r_s}$ provide precise boundary locations for the rough contour vertices and are used as anchor points of the final contour.
Since rough contours are spatially smoothed versions of the true boundaries, consecutive anchor points may not be adjacent along the raw contour. Therefore, the correct ordering of vertices in the raw contours is used to fill possible gaps between consecutive anchor points and to recover missed boundary segments.

Let $[{\bf C}_{q_s}^{\mathrm{raw}}]_{r_s}$ and $[{\bf C}_{q_{s+1}}^{\mathrm{raw}}]_{r_{s+1}}$ be two consecutive anchor points. Two cases are possible:

\begin{itemize}
    \item $q_{s+1} \neq q_s$: the closest points belong to different raw contours, indicating a transition between two boundary fragments. In this case, the two anchor points are directly connected in the final contour.

    \item $q_{s+1} = q_s$: both anchor points lie on the same raw contour.
    If $|r_{s+1}-r_s|$ is smaller than a predefined threshold, the two points belong to the same boundary sub-sequence and all intermediate raw vertices are inserted to preserve boundary continuity.
    The traversal direction along the raw contour is chosen according to the ordering of the anchor indices: if $r_{s+1}>r_s$, vertices are inserted in the forward cyclic order of ${\bf C}_{q_s}^{\mathrm{raw}}$, otherwise they are inserted in the reverse order, so that the final contour follows the cyclic orientation of the rough contour.
    Otherwise, a jump between two distant parts of the same raw contour is detected and the two anchor points are connected directly.
\end{itemize}

A hyperparameter $\mathrm{Jump\ Threshold}$ is used to distinguish between continuation along the same sub-sequence and a jump to another sub-sequence on the same raw contour.

\subsection{Thickness Calculation}\label{subsec:width}

From the previous section, we obtained the contours ${\bf{C}}_0,..., {\bf{C}}_m$ in the RGB image. Each RGB pixel has a corresponding 3D point in the point cloud captured by the camera, so we can represent each contour in 3D. This section addresses Problem~\ref{problem:widthcalculation}, where we calculate the thickness of the bio-printed material, which is crucial for evaluating the uniformity of the print.

We leverage the robot's recorded needle path
\[
Path = \{{\bf{p}}_0, ..., {\bf{p}}_n\},
\]
as the needle deposits material along this trajectory.
For each point ${\bf p}_l$ on the deposition path, we define the local thickness as the Euclidean distance between the two intersection points of the segmented bio-printed boundary with the plane passing through ${\bf p}_l$ and orthogonal to the local deposition path direction.
We assume printing occurs on a flat deposition plane and first estimate the best-fitting plane for the contours to define its normal vector. This plane allows us to identify which contour corresponds to a given path point and to determine the left and right side points for thickness measurement.

\paragraph{Fitting Plane}
Let ${\bf P}$ be the matrix of all vertices of ${\bf{C}}_0,..., {\bf{C}}_m$ with an additional column of ones. For a plane $ax+by+cz+d=0$ with $a^2+b^2+c^2+d^2=1$, minimizing the mean squared distance of points to the plane is equivalent to minimizing
\[
[a,b,c,d]{\bf P}^\top {\bf P} [a,b,c,d]^\top.
\]
The solution is the eigenvector of ${\bf P}^\top {\bf P}$ corresponding to its smallest eigenvalue. Let \({\bf v}\) be  the calculated plane normal.

\paragraph{Projecting Points and Contours}
To compute the thickness at ${\bf{p}}_l \in Path$, we project the contours and ${\bf{p}}_l$ onto the fitting plane, obtaining 2D polygons $\Tilde{{\bf{C}}}_0,..., \Tilde{{\bf{C}}}_m$. Using a ray-casting algorithm, we identify which polygon $\Tilde{{\bf{C}}}_{\Tilde{l}}$ contains the projected point. If none, thickness is zero; otherwise, ${\bf{p}}_l$ corresponds to ${\bf{C}}_{\Tilde{l}}$.

\paragraph{Thickness Measurement}
Let ${\bf t}_l$ be the tangent vector to the path at ${\bf{p}}_l$, approximated as
\[
{\bf t}_l \approx \frac{{\bf p}_{l+1}-{\bf p}_{l-1}}{||{\bf p}_{l+1}-{\bf p}_{l-1}||}.
\]
We define a plane through ${\bf{p}}_l$ with normal ${\bf t}_l$ and find its intersection with contour ${\bf{C}}_{\Tilde{l}}$. The closest intersection points on either side of ${\bf{p}}_l$ define the two side points. To separate points into left and right sets, we use the cross product ${\bf v} \times {\bf t}_l$: for an intersection point ${\bf q}$, if ${\bf{q}}-{\bf{p}}_l$ has acute angle with $\bf{v}\times {\bf{t}}_l$ i.e. $({\bf q}-{\bf p}_l) \cdot ({\bf v} \times {\bf t}_l) > 0$, then ${\bf q}\in {\bf L}$; otherwise, ${\bf q}\in {\bf R}$.

Finally, the thickness at ${\bf{p}}_l$ is
\[
\text{thickness}({\bf p}_l) = \Big\| \argmin_{{\bf q}\in {\bf L}} ||{\bf q}-{\bf p}_l||_2 - \argmin_{{\bf q}\in {\bf R}} ||{\bf q}-{\bf p}_l||_2 \Big\|_2.
\]

This procedure is repeated for all path points, producing a thickness profile of the bio-printed material along the deposited path.


\section{Evaluation Studies \& Results}\label{sec:experiments}

In this section, we evaluate our model using both 2D simulations and 3D real data, assessing performance in terms of accuracy and computational efficiency. In the first set of experiments, we perform 8 different cases of 2D simulations where the ground truth is available, allowing for both qualitative and quantitative evaluation. In the second set, we apply our algorithm to 4 cases of real data collected using our robotic setup and 3D camera, demonstrating its applicability to real-world scenarios.
For all studies, we set the $\mathrm{Jump\ Threshold} = 10$. The Canny detector, contour-finding, and dilation operations described in Sec.~\ref{sec:approach} are carried out using the Python OpenCV library \cite{itseez2015opencv}.

\subsection{Simulation Data}

\begin{figure*}[t]
\centering
\includegraphics[width=0.9\linewidth]{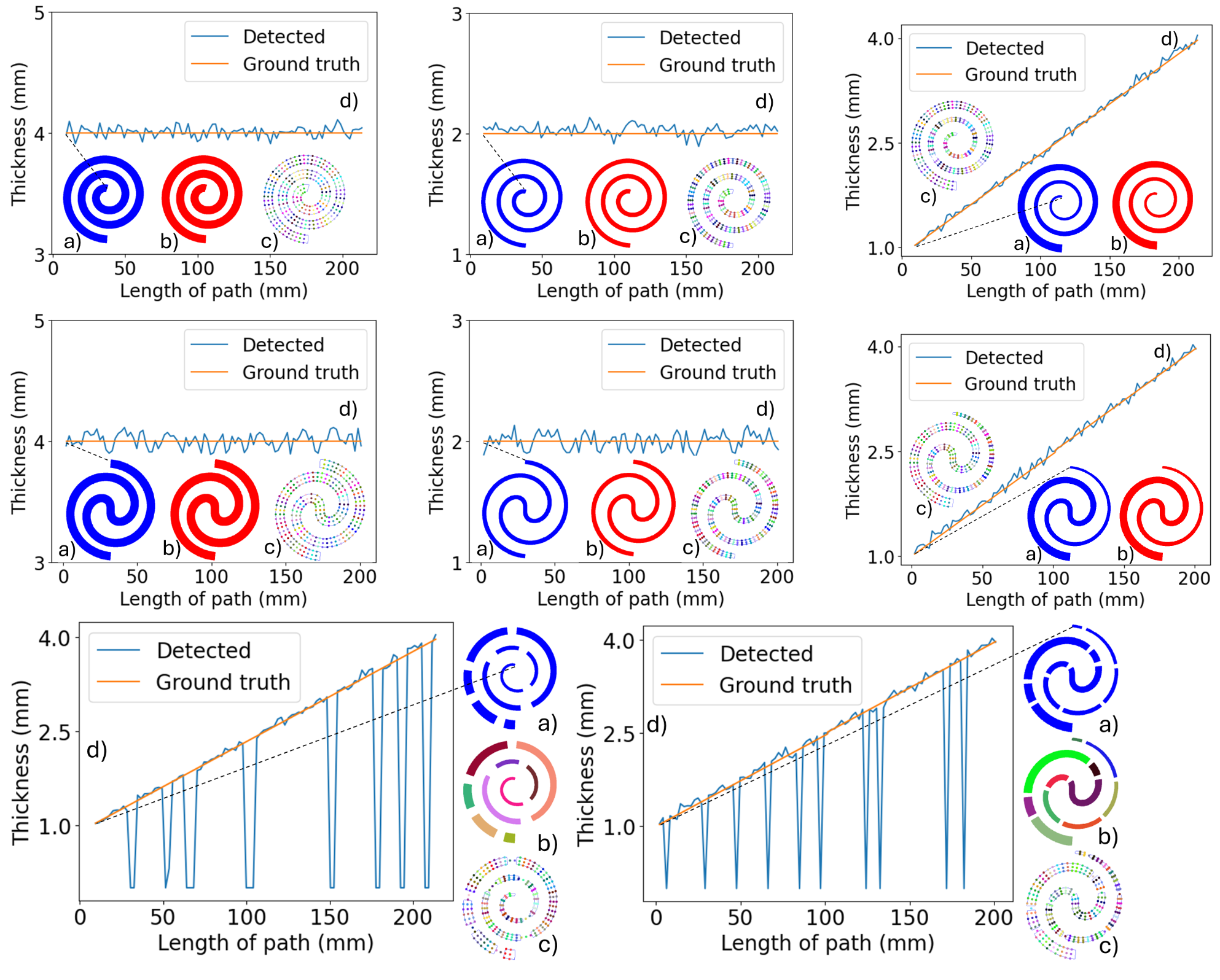}

 \caption{The 2D simulation samples generated by our simulation. All the Archimedean spirals are with angle $\pi\leq\phi\leq 6\pi$ and $a = 12$, and all of the Fermat Spirals are with maximum angle $\phi\leq 2\pi$ (for both branches) and $a=90$. In the first row, we see the Archimedean spirals from left with fixed thickness $4mm$ and, the second one with fixed thickness $2mm$ and the right one with linearly changing thickness from $1mm$ to $4mm$. The second row includes the Fermat spirals, from left with fixed thickness $4mm$ and the second one with fixed thickness $2mm$, and the right one with linearly changing thickness from $1mm$ to $4mm$. In the third row we see the disconnected spirals and linearly varying thickness from $1mm$ to $4mm$ (at the same time), with $10$ random gaps, each, with length about $2\%$ of the whole length of the curve (in the right one we have $10$ distinct gaps and on the left one we have two pairs of overlapped gaps). On the third row the left one is Archimedean and the right one is Fermat. In each figure, \textsf{a)} is the ground truth, \textsf{b)} is the areas detected by our algorithm filled with a random color. Then we uniformly (with respect to the length of the curve) pick about $100$ points from the path and find the side points for these path points. The distance between these side points is the thickness at that path point. In \textsf{c)}, each picked path point and its side points are shown with the same color, but the colors are random, the thickness along these $100$ points is shown in \textsf{d)}.}
\label{fig:2dsimulation}
\vspace{-5mm}
\end{figure*}

The 2D simulation cases include two complex spiral curves, Archimedean and Fermat, each with four different parameters as the ground truth. In this section, we provide a mathematical definition of the Archimedean and Fermat curves and present the simulation parameters. The simulation test cases are further explained.

\subsubsection{Archimedean spiral} In polar coordinates $(r, \phi)$, $r = a\phi$, with Cartesian parameterization $(x, y) = (a\phi\cos\phi, a\phi\sin\phi)$.

\subsubsection{Fermat spiral} In polar coordinates $(r, \phi)$, $r = \pm a\sqrt{\phi}$ ($\phi \geq 0$), with Cartesian coordinates $(x, y) = (a\sqrt{\phi}\cos\phi, a\sqrt{\phi}\sin\phi)$ for one branch, and $(-a\sqrt{\phi}\cos\phi, -a\sqrt{\phi}\sin\phi)$ for the other branch.

\subsubsection{Simulation parameters} $a = 12$ for Archimedean and $a = 90$ for Fermat spirals. As default, we set image dimensions to be $512 \times 512$ pixels, with each pixel representing $0.1 \mathrm{mm} \times 0.1 \mathrm{mm}$. Archimedean spirals span $\pi \leq \phi \leq 6\pi$, while Fermat spirals are limited to $\phi \leq 2\pi$ for both branches.
To simulate the robot path, $1000$ points are sampled uniformly along each curve. Uniform sampling ensures consistent arc-length spacing, matching the constant speed and fixed sampling rate of a real robotic deposition needle. In this regard, we use the formula of the length of each curve, then we use \texttt{scipy.optimize.fsolve} to find the corresponding $\phi$ of the uniform points on the curve.

\subsubsection{Thickness simulation} To synthetically create correct true thicknesses for each path point, we analytically compute the unit tangent vector, rotate it clockwise and counterclockwise, and move along these directions by $d/2$ to generate side points. Connecting side points creates polygons representing the BPC. Zero-thickness points are handled by segmenting consecutive zeros and applying the same procedure to the nonzero segments, resulting in disconnected polygons. Variable thickness and gaps are simulated by adjusting side points for each path point. Polygon vertices are then discretized to match image resolution, introducing minor discretization errors.

\subsubsection{Test cases} The evaluation studies include four different types of BPCs for both Archimedean and Fermat spirals, resulting in eight total test cases: (i) uniform thickness of $2$ mm, (ii) uniform thickness of $4$ mm, (iii) linearly varying thickness ($1$ mm to $4$ mm), (iv) disconnected BPCs with linearly varying thickness and a number of random gaps.
Without loss of generality, the number of random gaps were arbitrarily chosen to be $10$.
Each Gap's length is $\sim 2\%$ of the curve length and may overlap. To generate a disconnected shape, we uniformly (with respect to the curve arc length) sample 10 gap start points along the path. In Fig.~\ref{fig:2dsimulation}, in each image, \textsf{a)} shows the ground truth, \textsf{b)} the delineated BPC by our segmentation algorithm in Sec.~\ref{subsec:boundary}, \textsf{c)} sample path points with detected side points, and \textsf{d)} true and measured thickness along the path using our method in Sec.~\ref{subsec:width}.

\begin{figure*}[t]
\includegraphics[width=0.24\linewidth, trim={0mm 0mm 0mm 0mm}, clip]{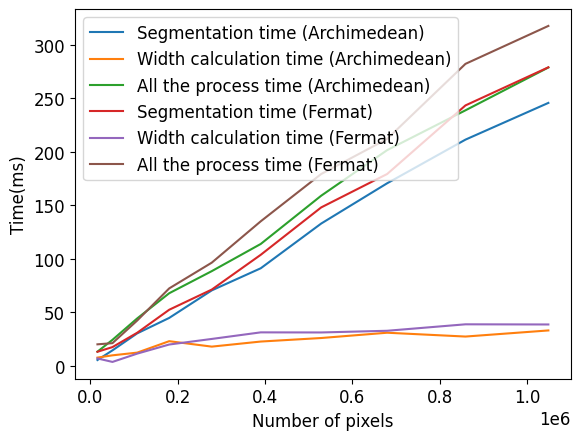}
\includegraphics[width=0.24\linewidth, trim={0mm 0mm 0mm 0mm}, clip]{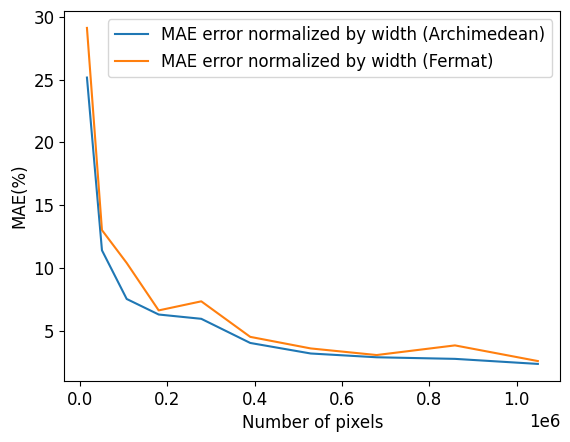}
\includegraphics[width=0.24\linewidth, trim={0mm 0mm 0mm 0mm}, clip]{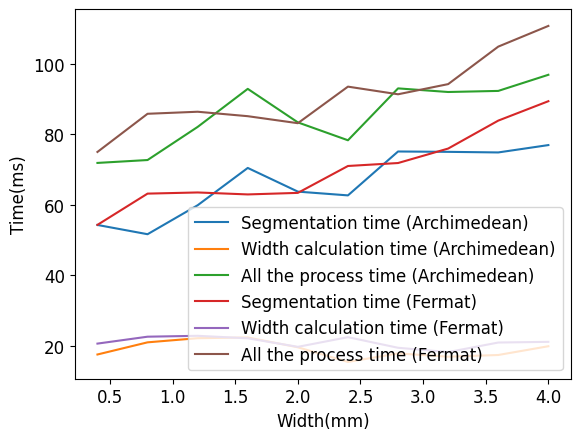}
\includegraphics[width=0.24\linewidth, trim={0mm 0mm 0mm 0mm}, clip]{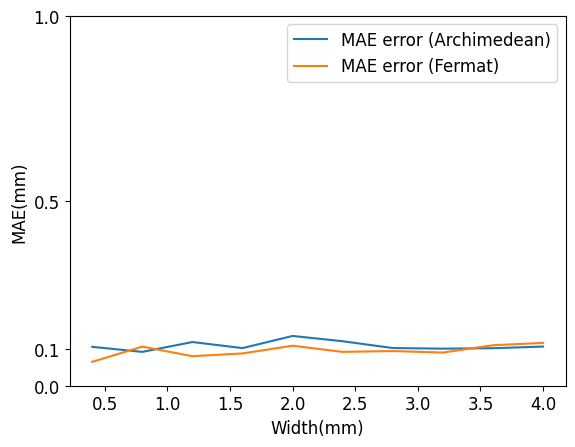}
\caption{From left to right, the first figure shows the effect of the resolution on the processing time of the algorithm. The second figure shows the effect of the resolution on the normalized MAE (in percentage). The third figure shows the effect of the thickness of the curve on the time of processing, and the last figure shows the effect of the thickness on the accuracy. For the first two left images, we pick $10$ equal distance samples as the ratio from $r = \frac{1}{4}$ to $r = 2$, we change the resolution through $\lfloor512 r\rfloor\times\lfloor512 r\rfloor$. The thickness of the curve stays $2mm$. For the right two images, we change the thickness for $10$ equal distance samples from $0.4mm$ to $4mm$. For all of the images for resolution or thickness, we generate $10$ curves with $10$ random gaps, then we get the mean of the target measurement (time or accuracy) for these $10$ random samples. The start point of gaps is uniformly picked, and the length of the gaps is about $2\%$ of the length of the whole curve.}
\label{fig:2dcurves}
\vspace{-2mm}
\end{figure*}

\begin{table}[t]
\centering
\caption{Simulation results. All units in mm. Curve types A and F represent Archimedean and Fermat curves respectively.}
\setlength{\tabcolsep}{3pt}
\begin{tabular}{|c|c|c|c|c|c|c|c|}
\hline
Case & Thickness & No. gaps & Curve type & MAE & Std. & Max & Min \\ \hline
Case 1 & 4   & 0  & A & 0.034 & 0.025 & 0.108 & 0.000 \\ \hline
Case 2 & 4   & 0  & F & 0.051 & 0.033 & 0.112 & 0.000 \\ \hline
Case 3 & 2   & 0  & A & 0.040 & 0.029 & 0.133 & 0.000 \\ \hline
Case 4 & 2   & 0  & F & 0.057 & 0.033 & 0.135 & 0.002 \\ \hline
Case 5 & 1--4 & 0  & A & 0.035 & 0.027 & 0.143 & 0.000 \\ \hline
Case 6 & 1--4 & 0  & F & 0.046 & 0.034 & 0.150 & 0.000 \\ \hline
Case 7 & 1--4 & 10 & A & 0.025 & 0.027 & 0.116 & 0.000 \\ \hline
Case 8 & 1--4 & 10 & F & 0.038 & 0.034 & 0.150 & 0.000 \\ \hline
\end{tabular}
\label{tab:2dsimulation}
\vspace{-4mm}
\end{table}

\subsection{Experimental Studies}

\begin{figure}[h]
\centering
\includegraphics[width=0.92\columnwidth]{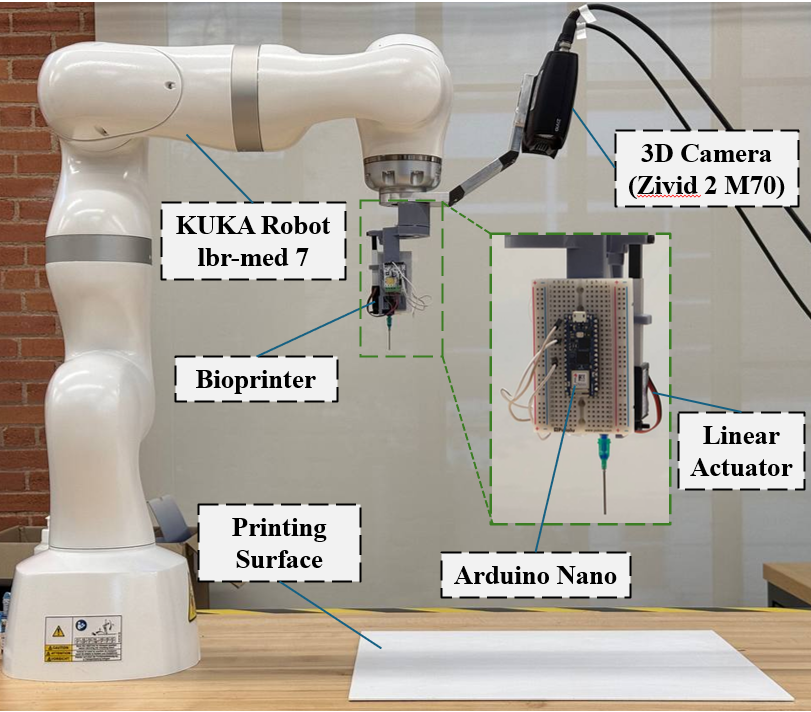}
\caption{The figure illustrates the experimental setup, which includes the KUKA robotic manipulator, Zivid 2 structure-light 3D camera, and the bioprinter tool containing the linear actuator, Arduino, a syringe containing the bioprinting material and a commercial needle as the nozzle.}
\label{fig:setup}
\end{figure}

For the experimental studies, we use a robotic bioprinting setup to print different geometries and acquire RGB and point cloud scans from the BPCs. Then, we apply our proposed edge detection and thickness calculation algorithms on the collected data.

The experimental robotic bioprinting setup as illustrated in Fig.~\ref{fig:setup}
includes a 7 degree of freedom robotic manipulator KUKA LBR Med 7 (Kuka, Germany) equipped with a custom made extruder-based bioprinter and a structured-light 3D camera (Zivid Two M70, Zivid, Norway) to take RGB and point cloud images from the printed filament. It is worth mentioning that the camera's point cloud resolution is $\approx$ 0.2 mm as reported by the manufacturer. The bioprinter is equipped with a linear actuator (Actuonix P8-50-165-3-ST linear stepper motor, Actuonix Motion Devices Inc., Canada), a Pololu Tic T834 motor driver (Pololu, USA), an Arduino Micro as the controller, a 50 ml commercial syringe for holding the gel, and a 18-gauge commercial needle serving as the nozzle \cite{Rezayof2024}.
Without loss of generality, a blue-dyed ultrasound gel (Aquasonic 100, Parker Laboratories Inc.) was used as the printing material.
Second generation of the Robot Operating System (ROS2, Humble) was used to control the robot \cite{Huber2024} and facilitate the communication with the Arduino (i.e. bioprinter's controller).

\begin{figure*}[t]
\centering
\includegraphics[width=0.86\linewidth]{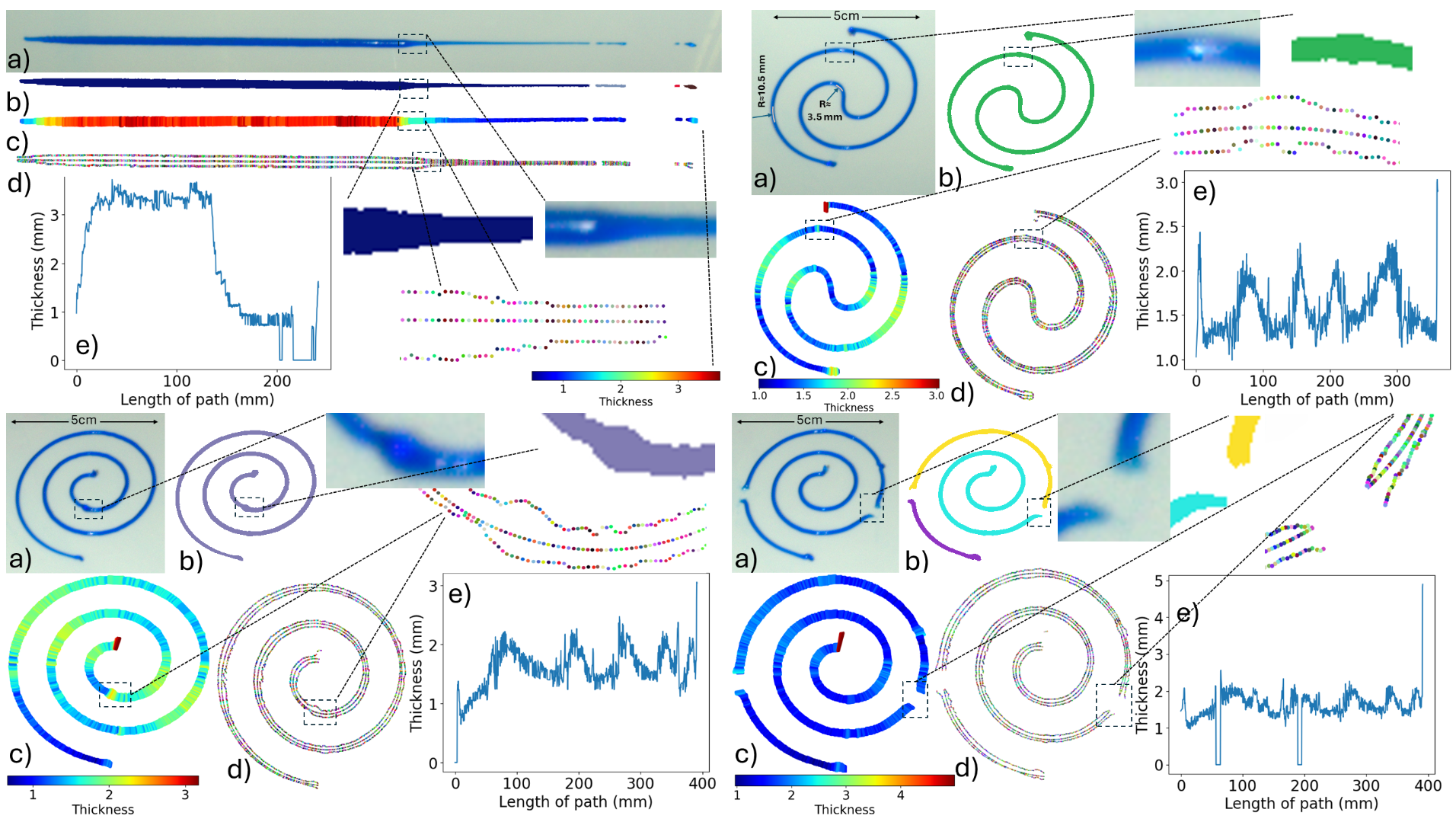}
\caption{Results of applying our algorithm to real data printed by a robotic system.
Top left: straight line with injection only along the first half of the path; Top right: Fermat curve.
Bottom left: Archimedean curve; Bottom right: Archimedean curve with gaps.
In each example, \textsf{a)} shows the ground-truth RGB image. After segmentation (Sec.~\ref{subsec:boundary}), the filled contour is shown in \textsf{b)}. The ordered contour pixels are mapped to the point cloud to reconstruct the 3D contour. Using Sec.~\ref{subsec:width}, we compute side points for each path point and define the thickness as the distance between them. Path points and their corresponding side points are shown in \textsf{d)}. The color map in \textsf{c)} indicates thickness, and \textsf{e)} plots thickness versus curve length.}
\label{fig:realdata}
\vspace{-4mm}
\end{figure*}

Four types of BPCs with different geometries were printed using the robotic bioprinting setup: (i) a straight line, (ii) a Fermat spiral curve, (iii) an Archimedean spiral curve and (iv) an Archimedean spiral curve with discontinuities.
The robot's linear velocity was set 3 mm/s constant while printing. In case 1, injection rate was set to be 0.144 mm/s at the beginning and set to 0 on the halfway. For the printed Fermat's and Archimedean spirals the injection rate was set at 0.027 mm/s and kept constant throughout the prints. For each case, the movement (path) of the needle, the RGB image and the point cloud taken by the camera were recorded to be later used in the proposed algorithms.
The results are provided in Fig.~\ref{fig:realdata}. For each case, we first acquired the RGB image which is shown as \textsf{a)}. Next, we processed the RGB image using our delineation and segmentation algorithms that is described in Sec.~\ref{subsec:boundary}, with a $3\times 3$ kernel for dilation. The results are shown in \textsf{b)}. In \textsf{b)} the detected contour is filled with color to clarify that the result is a closed well shaped contour that is able to distinguish the interior area from the exterior area of each bio-printing region. Then, we found the corresponding 3D contour. In each case, the path points and the side points are plotted in \textsf{d)}. Each path point and its side points are plotted with similar yet random colors.

\section{Discussion}
Figure~\ref{fig:2dsimulation} depicts the results of the 2D simulation studies. Investigating the figure indicates that the corresponding points on each boundary to the point on the needle path (i.e., the side points) are correctly detected, which their distance is the thickness of the BPC. The thickness versus length of needle path is provided in \textsf{d)}.
As the figure shows, it is visually seen that the side points align with intuition, and measured thickness matches ground truth with sub-millimeter ($41\quad\mu$m on average, see Table~\ref{tab:2dsimulation}) accuracy. In disconnected cases, zero-thickness regions are correctly identified. For example,  in case \textsf{c)}, only the path point is shown (instead of a path point and two side points), and the corresponding zero values are observed in case \textsf{d)} as gap regions.
The numerical results for the simulation studies are provided in Table~\ref{tab:2dsimulation}. This table presents the true thickness of the simulated BPC, number of gaps (if any), curve type (A: Archimedean and F: Fermat curves), the Mean Absolute Error (MAE), standard deviation, and the maximum and minimum of the absolute error values.
As it can be seen in the table, all the MAEs were less than $0.06$ mm (i.e., $60\quad\mu$m) showing the exceptional accuracy of the proposed thickness measurement method.
With only $0.057$ mm of error and a maximum error value of $0.135$ mm, Case 4 held the maximum average error between all cases.
Furthermore, the maximum value over different cases of the maximum error was $150\quad\mu$m.
This maximum error $\approx 0.15$ mm corresponds to the size of one pixel, indicating discretization as the main error source in the thickness measurement algorithms.

Furthermore, Fig.~\ref{fig:2dcurves} illustrates algorithm performance under varying resolutions and thicknesses.
For resolution analysis, we set the thickness to $2$ mm and vary the image resolution using $10$ ratios in the range $r=\frac{1}{4}$ to $2$, resulting in images of size $\lfloor512 r\rfloor\times\lfloor512 r\rfloor$.
Accordingly, the physical size of each pixel varies as $\frac{0.1}{r}$ mm $\times$ $\frac{0.1}{r}$ mm.
For each resolution, we generated 10 disconnected spirals, each with 10 random gaps, and repeat the experiment 10 times.
The thickness calculation MAE and the computation time of segmentation, thickness estimation, and the overall pipeline are averaged over 10 runs.
As can be observed, time increases roughly linearly with the number of pixels, while the accuracy degrades rapidly as the resolution decreases.
For thickness analysis, BPC thickness varies from $0.4$ mm to $4$ mm. Similarly, for each thickness, we generated $10$ disconnected curves with $10$ random gaps. Then, we measured the times and MAE errors.
As it is seen in Fig.~\ref{fig:2dcurves}, the accuracy remains largely independent of thickness, while processing time slightly increases with thicker BPCs. Specifically, the total time (including the segmentation and the thickness calculation) increased from $\approx70$ ms for a thickness of $0.5$ mm to $\approx90$ ms for a thickness of $4.0$ mm. This increment is mainly due to discretization effects, since thicker constructs cover more pixels and produce more detailed boundary polygons.

\begin{table}[t]
\centering
\caption{Results from hardware experiments. All units in mm.}
\setlength{\tabcolsep}{4pt}

\begin{tabular}{|c|c|c|c|c|c|c|}
\hline
Case & Pattern & Avg. Thickness & Std. & Q1 & Q2 & Q3 \\ \hline
Case 1 & Line & 2.135 & 1.280 & 0.917 & 2.790 & 3.322 \\ \hline
Case 2 & Fermat & 1.569 & 0.289 & 1.352 & 1.479 & 1.769 \\ \hline
Case 3 & Archimedean & 1.604 & 0.330 & 1.428 & 1.593 & 1.847 \\ \hline
Case 4 & \begin{tabular}[c]{@{}c@{}}Archimedean \\ (w/ gaps)\end{tabular} & 1.575 & 0.411 & 1.424 & 1.586 & 1.827 \\ \hline
\end{tabular}
\label{tab:realdata}
\end{table}

The results of the experimental studies are presented in Fig.~\ref{fig:realdata} and Table~\ref{tab:realdata}. As there are no ground truth values for the thickness in the experimental studies, the table reports the mean, standard deviation, and different quartiles of the thickness over the print length. Since the calculated thickness may contain linearization noise, we include quartiles instead of min and max.
Of note, case 1 includes a higher range in its std ($1.280$ mm) as the thickness was decaying when the injection was turned off halfway through the print. Nevertheless, the analysis of the Fermat and the Archimedean spirals show lower std values ($0.34$ mm on average) showing that their thickness was more uniform. Overall, these results indicate the high performance of the proposed vision-based approach in automating the thickness evaluation step in a realistic bioprinting procedure.

\section{Conclusion}
In this paper, leveraging point cloud scans and RGB images obtained from a structured-light 3D camera, we proposed a fully automated and highly accurate vision-based method for evaluating the thickness of BPCs in a robotic bioprinting procedure.
This quantitative measurement is a key prerequisite for reliable assessment of printing quality in bioprinting applications.
We proposed novel segmentation methods for masking the BPCs from the background using the RGB images, and introduced geometry-driven thickness measurement algorithms that enable quantitative evaluation of the BPCs' thickness based on the captured point cloud scans.
The proposed methods were validated using both simulation and experimental evaluation studies.
In the 2D simulation studies the proposed framework achieved mean absolute thickness errors between $0.025$\,mm and $0.057$\,mm over a wide range of shapes, thickness profiles, and disconnected cases, with a spatial resolution of $0.1\,\mathrm{mm}\times0.1\,\mathrm{mm}$ per pixel.
In the future, we will integrate the proposed framework into closed-loop bioprinting procedures with our robotic system, where on-demand feedback is required to dynamically modulate printing parameters such as deposition rate and robot velocity, and multi-layer multi-material bioprinting tasks.

\bibliographystyle{ieeetr}
\bibliography{root.bib}

\end{document}